\documentclass[10pt,twocolumn,letterpaper]{article}

\usepackage[pagenumbers]{cvpr} 

\usepackage{graphicx}
\usepackage{amsmath}
\usepackage{amssymb}
\usepackage{booktabs}
\usepackage{color}
\usepackage{nicefrac}

\usepackage{etoolbox}
\makeatletter
\pretocmd{\chapter}{\addtocontents{toc}{\protect\addvspace{15\p@}}}{}{}
\pretocmd{\section}{\addtocontents{toc}{\protect\addvspace{5\p@}}}{}{}
\pretocmd{\subsection}{\addtocontents{toc}{\protect\addvspace{3\p@}}}{}{}
\makeatother
\usepackage{titletoc}

\usepackage{xpatch}
\usepackage{array}
\usepackage{multirow}
\usepackage{graphicx}
\usepackage{xcolor,colortbl}
\usepackage{pifont}
\usepackage{enumitem}
\usepackage{xcolor}
\PassOptionsToPackage{dvipsnames}{xcolor}

\definecolor{Gray}{gray}{0.9}

\definecolor{indian red}{RGB}{205,92,92}

\usepackage{comment}
\usepackage[accsupp]{axessibility}

\definecolor{mycitecolor}{rgb}{0, 0.4, 0.7}
\usepackage[pagebackref,breaklinks,colorlinks,bookmarks,citecolor=mycitecolor]{hyperref}

\usepackage[capitalize]{cleveref}
\crefname{section}{Sec.}{Secs.}
\Crefname{section}{Section}{Sections}
\Crefname{table}{Table}{Tables}
\crefname{table}{Tab.}{Tabs.}

\def\cvprPaperID{1655} 
\def\confName{CVPR}
\def\confYear{2025}

\begin{document}

\title{Local-Global Attention: An Adaptive Mechanism for Multi-Scale Feature Integration}

\author{Yifan Shao \\
Lanzhou University of Technology\\
{\tt\small 210031501038@lut.edu.cn}}

\maketitle
\begin{abstract}

In recent years, attention mechanisms have significantly enhanced the performance of object detection by focusing on key feature information. However, prevalent methods still encounter difficulties in effectively balancing local and global features. This imbalance hampers their ability to capture both fine-grained details and broader contextual information—two critical elements for achieving accurate object detection.
To address these challenges, we propose a novel attention mechanism, termed Local-Global Attention, which is designed to better integrate both local and global contextual features. Specifically, our approach combines multi-scale convolutions with positional encoding, enabling the model to focus on local details while concurrently considering the broader global context. Additionally, we introduce learnable $\alpha$ parameters, which allow the model to dynamically adjust the relative importance of local and global attention, depending on the specific requirements of the task, thereby optimizing feature representations across multiple scales.
We have thoroughly evaluated the Local-Global Attention mechanism on several widely used object detection and classification datasets. Our experimental results demonstrate that this approach significantly enhances the detection of objects at various scales, with particularly strong performance on multi-class and small object detection tasks. In comparison to existing attention mechanisms, Local-Global Attention consistently outperforms them across several key metrics, all while maintaining computational efficiency. Code is available at the \href{https://github.com/ziyueqingwan/LocalGlobalAttention}{link}.

\end{abstract}
    
\section{Introduction}
\label{sec:intro}

In recent years, significant progress has been made in the field of object detection, with many methods achieving remarkable improvements in performance metrics \cite{girshick2015fast, liu2016ssd, Redmon_2016_CVPR}. Nevertheless, researchers continue to explore new approaches to further enhance accuracy and efficiency, especially in challenging scenarios such as multi-class and small object detection. Attention mechanisms have emerged as an effective means to improve model performance \cite{vaswani2017attention, hu2018squeeze, wang2020eca, woo2018cbam, zhang2021sa}, gaining considerable attention in deep learning due to their ability to significantly boost performance with only a limited increase in computational cost.

Among widely-used attention mechanisms, local and global attention are particularly noteworthy \cite{luong2015effective}. Local attention focuses on fine-grained, localized details within the input, capturing essential local information \cite{yang2016hierarchical}. In contrast, global attention emphasizes the overall content and global context of the input, which is crucial for understanding broader relationships and larger patterns \cite{bahdanau2014neural, Redmon_2016_CVPR}. Despite their strengths, both types of attention have inherent limitations: local attention often overlooks global dependencies, while global attention sacrifices detailed local information. Although many researchers have attempted to combine local and global features, these efforts often face challenges in effectively balancing the two, resulting in either suboptimal model performance or significant computational overhead that offsets performance gains \cite{wu2019fbnet, Zhang_2022_CVPR, liu2021swin, yang2021focal}.

In this paper, we propose a novel attention mechanism called Local-Global Attention, designed to balance local and global features by integrating multi-scale convolution and positional encoding. This enables the model to capture both local details and global context. Additionally, we introduce learnable $\alpha$ parameters, allowing the model to dynamically adjust the balance between local and global attention in a data-driven manner, achieving consistent improvements across various datasets. Overall, Local-Global Attention ensures optimized feature representation across different scales, enhancing detection performance while maintaining low computational costs.

Specifically, the implementation of the Local-Global Attention mechanism is as follows: After initial feature extraction, we apply multi-scale convolutions with smaller kernels to capture localized features, which are then aggregated to form a local attention map. In parallel, larger kernel convolutions combined with positional encoding are used to extract broader global features, generating a global attention map. These two attention maps are then fused using learnable $\alpha$ parameters, creating a unified attention map, which is subsequently passed through additional network layers for further refinement and processing.

\begin{figure}
    \centering
    \includegraphics[width=1\linewidth]{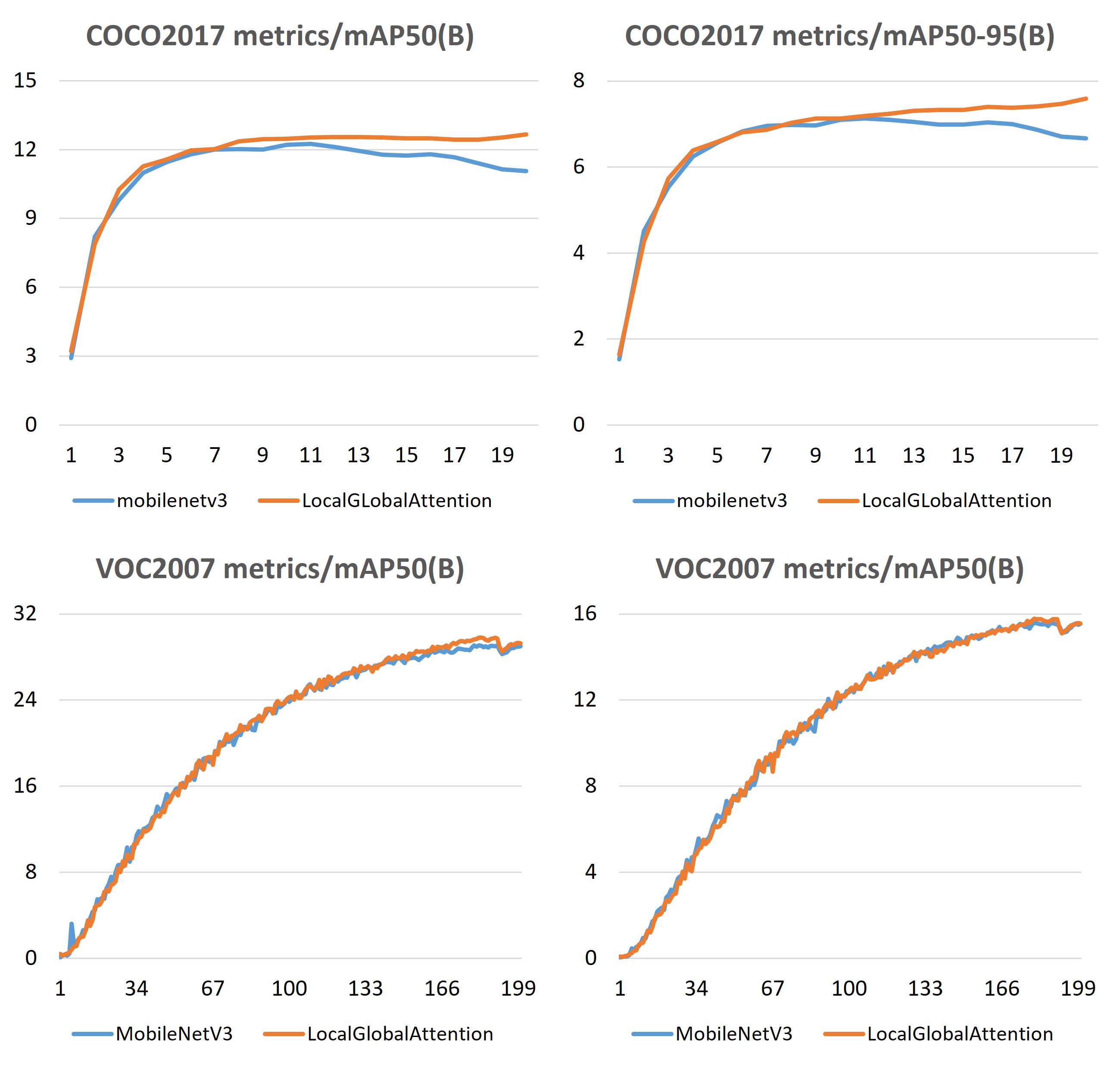}
    \caption{The mAP@50 and mAP@50:95 results on the COCO2017 \cite{lin2015microsoft} and VOC2007 \cite{pascal-voc-2007} datasets compare the performance of MobileNetV3 \cite{howard2019searching} with its enhanced version using the Local-Global Attention mechanism. On COCO2017 \cite{lin2015microsoft}, all models were trained for 20 epochs using the Adam optimizer, while on VOC2007 \cite{pascal-voc-2007}, training was conducted for 200 epochs with the AdamW optimizer. In both cases, other settings followed the YOLOv8 \cite{yolov8_ultralytics} default configuration.}
    \label{fig1}
\end{figure}

This approach provides several key advantages. Firstly, it captures both fine-grained and large-scale information, offering a more comprehensive understanding of the input, which enables more accurate localization and identification of targets. Secondly, the mechanism is lightweight and modular, making it easy to integrate into a variety of existing network architectures, such as MobileNetV3 \cite{howard2019searching} and ResNet \cite{he2016deep}. By prioritizing informative features and filtering out irrelevant data, Local-Global Attention significantly enhances detection accuracy in detection tasks, as shown in Figure \ref{fig1}.

To validate the effectiveness of our method, we conducted extensive experiments on multiple benchmark datasets, including VOC2007 \cite{pascal-voc-2007}, VOC2012 \cite{pascal-voc-2012}, VisDrone2019-DET \cite{zhu2021detection}, TinyPerson \cite{yu2020scale}, COCO2017 \cite{lin2015microsoft}, GWHD2020 \cite{david2020global}, COCO minitrain \cite{HoughNet}, DOTA-v1.0 \cite{Xia_2018_CVPR}, as well as MNIST \cite{deng2012mnist} and Fashion-MNIST \cite{xiao2017/online}. The results demonstrate that Local-Global Attention outperforms existing attention mechanisms with similar computational requirements, consistently improving the model’s detection accuracy.

In conclusion, we believe that Local-Global Attention is a practical and efficient solution that addresses some limitations of existing attention mechanisms, offering a flexible approach that moves closer to more accurate and computationally feasible object detection models.

\section{Related Work} \label{sec}

In recent years, attention mechanisms have gained increasing attention due to their ability to selectively focus on the most relevant features within data, significantly improving the performance of various computer vision tasks. This section reviews the development and applications of attention mechanisms.


\subsection{YOLO}

YOLOv8 \cite{yolov8_ultralytics}, as a classic model in the YOLO series \cite{redmon2018yolov3, bochkovskiy2020yolov4, yolov5, li2023yolov6, wang2022yolov7, yolov8_ultralytics}, builds upon previous YOLO versions with advancements in the backbone and neck architectures, anchor-free split Ultralytics head, and optimizations in speed and accuracy. It supports a wide range of computer vision tasks, including object detection, instance segmentation, pose/keypoint detection, oriented object detection, and classification.

\subsection{Neural Networks}

The MobileNet architecture family \cite{howard2017mobilenets, sandler2018mobilenetv2, howard2019searching} has made significant progress in lightweight neural networks, particularly MobileNetV3 \cite{howard2019searching}, which is a representative of lightweight neural networks, especially suited for mobile hardware. It combines Neural Architecture Search (NAS) with manually designed elements, utilizing inverted residual blocks to reduce memory usage and computational demands. Additionally, it integrates the Squeeze-and-Excitation attention \cite{hu2018squeeze} module, further optimizing feature representations and enhancing multitask performance.

The ResNet series, especially ResNet18 \cite{he2016deep}, addresses the vanishing gradient problem by introducing residual learning, making it possible to construct very deep networks without gradient issues. The residual connections enhance gradient flow, promoting the effective integration of low-level and high-level features, making it a robust backbone model for a variety of applications.

The backbone of YOLOv8 \cite{yolov8_ultralytics} is similar to CSPDarkNet53 \cite{bochkovskiy2020yolov4}, performing exceptionally well in real-time object detection. It efficiently extracts both spatial and semantic information from the input data, achieving a good balance between speed and accuracy, particularly suited for mobile and edge devices.


\subsection{Attention Mechanisms}

Attention mechanisms have significantly impacted many computer vision tasks, resulting in various methods aimed at improving feature representation. Below are some commonly used attention mechanisms and their primary strengths and limitations.

Multi-head Self-Attention \cite{vaswani2017attention} has become a mainstay for capturing long-range dependencies across different parts of an input. By using multiple attention heads in parallel, it can focus on various aspects of the input data. While highly effective for modeling global context, it often struggles with capturing fine-grained local details, which is crucial for tasks requiring both local precision and global awareness.

queeze-and-Excitation Attention \cite{hu2018squeeze} enhances channel-wise feature representation by dynamically recalibrating channel weights, improving the model’s ability to focus on essential features.

Convolutional Block Attention Module \cite{woo2018cbam} extends attention mechanisms by sequentially applying channel and spatial attention, refining feature maps step by step.

While these attention mechanisms each have unique advantages, they also face challenges when it comes to integrating both local and global features in complex scenarios. To address these challenges, our work builds on these methods by proposing the Local-Global Attention mechanism, which combines multi-scale convolution and positional encoding for a more effective integration of local and global information.

\section{Local-Global Attention Mechanism}

\begin{figure}
    \centering
    \includegraphics[width=0.95\linewidth]{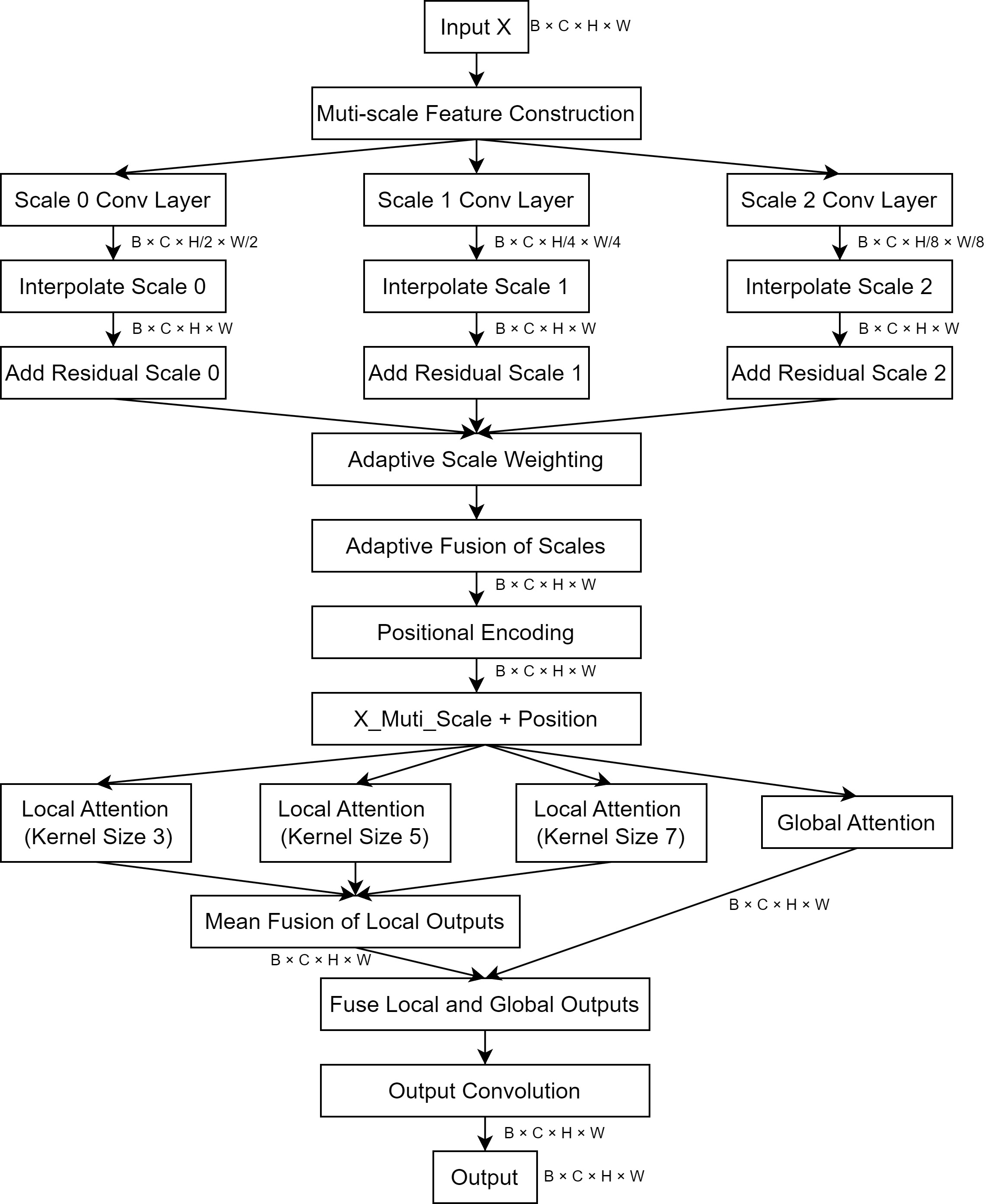}
    \caption{Local-global attention structure diagram}
    \label{fig3}
\end{figure}

The local-global attention mechanism aims to optimize feature extraction by integrating both local and global contextual information. This balanced strategy helps the model focus on fine details in specific regions while capturing broad information from the input data, providing a comprehensive understanding of both local and global contexts. The structure of the local-global attention mechanism is shown in Figure \ref{fig3}. In the following, we will describe each component of the local-global attention mechanism in detail.

\subsection{Input Representation}
The structure of the input tensor \( \mathbf{X} \in \mathbb{R}^{B \times D \times H \times W} \) includes:
\begin{itemize}
    \item \( B \): Batch size, representing the number of samples processed in parallel.
    \item \( D \): Embedding dimension, indicating the depth of the features, providing rich descriptions for each spatial position.
    \item \( H \): Spatial dimension, representing the height of the feature map.
    \item \( W \): Spatial dimension, representing the width of the feature map.
\end{itemize}
The structure of the input tensor preserves both spatial and depth information, allowing the model to efficiently capture spatial relationships between different regions in the image or feature map \cite{hu2018squeeze, yang2021focal, dosovitskiy2020image}. Through the design of the depth dimension, multi-dimensional and multi-level feature representations are provided for each spatial position, thereby better handling the complex patterns and contextual information in the image.

\subsection{Head Dimension and Scaling}
To ensure the stability and efficiency of the attention calculation, we divide the embedding dimension \( D \) into multiple heads, with each head focusing on a different subset of the features:
\begin{equation}
    \text{head\_dim} = \frac{D}{\text{num\_heads}}
\end{equation}
Each head processes a different part of the data, introducing diverse representations and capturing different aspects of the input.

To stabilize the gradients, a scaling factor is applied to normalize the attention scores:
\begin{equation}
    \text{scale} = \text{head\_dim}^{-0.5}
\end{equation}

\subsection{Multi-Scale Feature Extraction}
To better capture information at different resolutions, we use multi-scale convolutional layers with residual connections to extract features from both local and broader contexts. The original input \( \mathbf{X} \) is included as a baseline, ensuring that the model can access unchanged information during multi-scale interpretation.

\textbf{Multi-scale Convolutional Layers.}
For each scale \( i \), a specific set of convolutions is applied to capture features at a particular granularity:
\begin{equation}
    \mathbf{Y}_i = f_{\text{conv}}^{(i)}(\mathbf{X}) + \mathbf{X}
\end{equation}
where \( f_{\text{conv}}^{(i)} \) is a combination of depthwise convolutions and \(1 \times 1\) convolutions. Depthwise convolutions isolate each channel, effectively capturing spatial patterns, while \(1 \times 1\) convolutions integrate cross-channel information. The residual connection with the input retains important information and improves gradient flow, aiding in deeper and more efficient feature extraction.

The multi-scale feature set, including the original input, is:
\begin{equation}
    \mathbf{X}_{\text{multi-scale}} = \left\{ \mathbf{X} \right\} \cup \left\{ \mathbf{Y}_i \right\}_{i=1}^{\text{num\_scales}}
\end{equation}
This arrangement preserves both local details and broader context, enhancing the network's flexibility when handling diverse data patterns.

\subsection{Positional Encoding}
The positional encoding \( \text{PE} \in \mathbb{R}^{1 \times D \times H' \times W'} \) aligns with the spatial dimensions of the input feature map. This encoding is adjusted as needed and combined with the features, enabling the network to maintain the spatial relationships between elements. This is particularly useful for tasks that are sensitive to the relative positions of features, such as object detection and segmentation.

\subsection{Adaptive Scale Weights}
To adaptively emphasize features from different scales, we learn the weights \( \alpha_i \):
\begin{equation}
    \boldsymbol{\alpha} = \text{Softmax}(\mathbf{W}_{\text{scale}} \ast \mathbf{X})
\end{equation}
where \( \boldsymbol{\alpha} \in \mathbb{R}^{B \times (\text{num\_scales} + 1) \times H \times W} \). The Softmax operation normalizes these weights, forming a probability distribution across scales.

Applying these adaptive weights, the multi-scale features are combined into a single, refined feature map:
\begin{equation}
    \mathbf{X}_{\text{weighted}} = \sum_{i=0}^{\text{num\_scales}} \alpha_i \cdot \mathbf{X}_{\text{multi-scale}}[i]
\end{equation}
The weighted sum enables the model to dynamically focus on the relevant scales based on the input, optimizing its response to different feature patterns.

\subsection{Attention Mechanism}
The local-global attention mechanism simultaneously employs both local and global attention, efficiently capturing contextual details across different scales. This dual attention approach allows the model to focus on both the details in the input data and the broader structure, effectively bridging the gap between local features and global information.

\subsubsection{Local Attention}
To process local patterns at different granularities, local attention applies convolutions with varying kernel sizes \( k \in \{3, 5, 7\} \):
\begin{equation}
    Q_k, K_k, V_k = f_{\text{attn}}^{(k)}(\mathbf{X}_{\text{weighted}} + \text{PE})
\end{equation}
Using multiple kernel sizes allows the model to capture local features ranging from fine to more general, making it particularly effective in recognizing targets and textures with varying levels of complexity.

The local attention scores are computed as:
\begin{equation}
    \text{energy}_k = \frac{Q_k K_k^\top}{\sqrt{\text{head\_dim}}}
\end{equation}
The scaling factor stabilizes the gradient flow and balances the feature dimensions, aiding the model in converging more efficiently.

The attention weights are then obtained:
\begin{equation}
    \text{attention}_k = \text{Softmax}(\text{energy}_k)
\end{equation}
This emphasizes the relevant features based on learned relationships, allowing each kernel size to focus on a unique subset of spatial information.

The output for each kernel size is:
\begin{equation}
    \text{local\_out}_k = \text{attention}_k \cdot V_k + \mathbf{X}
\end{equation}
The residual connection ensures that the model integrates the newly attended features while retaining the original information, enhancing the details while maintaining a solid baseline.

The final local output is obtained by averaging across different kernel sizes:
\begin{equation}
    \text{local\_out} = \frac{1}{N} \sum_{k} \text{local\_out}_k
\end{equation}
This allows the model to fuse fine-grained local information from multiple perspectives, forming a comprehensive local feature set that covers a wide range of visual elements.

\subsubsection{Global Attention}
Global attention uses a larger convolution kernel \( k_g \), enabling the model to capture global spatial patterns in the input:
\begin{equation}
    Q_{\text{global}}, K_{\text{global}}, V_{\text{global}} = f_{\text{attn}}^{(k_g)}(\mathbf{X}_{\text{weighted}} + \text{PE})
\end{equation}
The larger kernel captures broader dependencies in the input, making this attention particularly suited for tasks requiring understanding of larger structures or long-range relationships.

The global attention output is:
\begin{equation}
    \text{global\_out} = \text{Softmax}\left(\frac{Q_{\text{global}} K_{\text{global}}^\top}{\sqrt{\text{head\_dim}}}\right) V_{\text{global}} + \mathbf{X}
\end{equation}
By utilizing a global perspective, this helps the model effectively gather high-level contextual information, complementing the fine-grained details captured by local attention, and facilitating a balanced understanding of both micro and macro structures.

\subsection{Output Fusion}
To integrate the outputs of local and global attention, we introduce learnable weight parameters \( \alpha_{\text{local}} \) and \( \alpha_{\text{global}} \) as weight factors:
\begin{equation}
    \text{out} = \alpha_{\text{local}} \cdot \text{local\_out} + \alpha_{\text{global}} \cdot \text{global\_out}
\end{equation}
The learnable weight parameters enable the model to dynamically adjust the importance of local or global features, allowing the network to focus according to task or data characteristics, thus providing responsive and context-aware feature representations. The combination of local and global features ensures that the model can adaptively balance the details and broader context of each input, forming a detailed and multi-dimensional feature set.

\subsection{Final Convolution and Output}
To further refine and compress the combined features, a \( 1 \times 1 \) convolution is applied:
\begin{equation}
    \text{final\_output} = f_{\text{conv}}^{(1 \times 1)}(\text{out})
\end{equation}
This operation helps to compress the rich information from both local and global attention into a more compact and efficient representation. The final output is optimized for downstream tasks such as classification or detection, providing both spatial details and context-aware information within a single feature map. Including this \( 1 \times 1 \) convolution also helps reduce dimensions without sacrificing key information. By merging features at this stage, the network ensures minimal computational overhead while maintaining a high representation capability, making it a practical choice for applications where both accuracy and efficiency are crucial.

\section{Experiments}

\subsection{Datasets}
 
\textbf{VOC2012} dataset primarily aims to identify objects from multiple visual categories in real-world scenes (i.e., objects that are not pre-segmented). It includes a total of 11,530 images in the training and testing sets across 20 categories, with 27,450 annotated ROI objects and 6,929 segmentation annotations.

\textbf{VisDrone2019} dataset consists of 288 video clips, comprising 261,908 frames and 10,209 static images captured by various drone cameras across different scenes, weather, and lighting conditions. Over 2.6 million frequently observed objects (such as pedestrians, cars, bicycles, and tricycles) have been manually annotated with bounding boxes. Additionally, important attributes such as scene visibility, object class, and occlusion are provided to facilitate better data utilization.

\textbf{TinyPerson} dataset is the first benchmark focused on detecting small objects (people) at a distance within large background contexts, containing 1,610 labeled images and 72,651 object annotations with bounding boxes.

\textbf{DOTAv1.0} dataset includes 15 common object categories such as ships, vehicles, and buildings, with 2,806 high-resolution aerial images and 188,282 instances. The image sizes range from 800 × 800 to 20,000 × 20,000 pixels. This dataset provides detailed bounding box annotations and is widely used to evaluate model performance on object detection tasks in aerial images, especially suitable for handling objects with significant variations in scale and orientation.

\textbf{COCOminitrain} dataset is a selected mini-train set from COCO2017 \cite{lin2015microsoft}, containing 25K images, 80 object categories, and approximately 184K annotations, accounting for about 20\% of train2017. It is particularly useful for hyperparameter tuning and reducing the cost of ablation experiments. The object instance statistics in COCOminitrain closely match those in train2017, and model performance on val2017 trained on MiniTrain is strongly positively correlated with that of the same model trained on train2017’s full dataset.

\textbf{GWHD 2020} dataset is the first large-scale dataset for detecting wheat heads from in-field optical images, covering a wide range of wheat varieties from different continents. It includes 4,700 high-resolution RGB images with 190,000 annotated wheat heads. These images were collected from several countries, covering wheat at different growth stages and with diverse genotypes. The images are annotated with wheat head bounding boxes to evaluate trained models' performance on unseen genotypes, environments, and observation conditions.

\textbf{MNIST} dataset is a classic dataset for handwritten digit recognition, containing 70,000 grayscale images, of which 60,000 are for training and 10,000 for testing. Each image is a 28x28-pixel image of handwritten digits ranging from 0 to 9.

\textbf{Fashion-MNIST} dataset is a more challenging fashion dataset intended to serve as a replacement for MNIST \cite{deng2012mnist}, containing 70,000 grayscale images of fashion items across 10 categories (e.g., T-shirts, shoes). The training set includes 60,000 examples, and the test set contains 10,000 examples, with each image sized at 28x28 pixels.

\subsection{Experimental Setup}
All datasets were trained according to YOLOv8 standards \cite{yolov8_ultralytics}. For the MNIST \cite{deng2012mnist} and Fashion-MNIST \cite{xiao2017/online} datasets, training was conducted over 20 epochs. TinyPerson \cite{yu2020scale} and GWHD2020 \cite{david2020global} datasets were trained over 200 epochs, while VOC2012 \cite{pascal-voc-2012}, VisDrone2019 \cite{zhu2021detection}, COCOminitrain \cite{HoughNet}, and DOTAv1.0 \cite{Xia_2018_CVPR} datasets were trained for 100 epochs. Training was performed using a GPU 4090, with training parameters kept consistent with YOLOv8 default settings \cite{yolov8_ultralytics}.

\subsection{Classification Experiments}

We integrated various attention mechanisms into MobileNetV3 to evaluate classification performance on the MNIST and Fashion-MNIST datasets. The experiments were based on YOLOv8's default parameter settings, and performance of each attention mechanism was assessed through Top-1 and Top-5 accuracy (\%) as well as computational complexity (GFLOPs).

\textbf{MNIST Dataset Experiment}: We integrated several attention mechanisms (MHSA, SE, CBAM, and our proposed Local-Global Attention) into the MobileNetV3 model and compared classification performance on the MNIST dataset. Table \ref{tabel1} presents the Top-1 and Top-5 accuracy and GFLOPs for each attention mechanism. The results demonstrate that the model with Local-Global Attention achieved the highest Top-1 accuracy of 99.4\%, further enhancing classification accuracy without increasing computational complexity.

\begin{table}[!ht]
    \centering 
    \caption{Classification performance of different attention mechanisms integrated into the MobileNetV3 model on the MNIST dataset \cite{deng2012mnist}.} \label{tabel1} 
    \resizebox{\linewidth}{!}{
    \begin{tabular}{lcccc}
    \hline
        Backbone & Top-1 Acc (\%) & Top-5 Acc (\%) & GFLOPs  \\ \hline 
        MobileNetV3 & 99.3 & 100   & 0.9  \\ 
        +MHSA & 99.3 & 100   & 0.9  \\ 
        +SE & 99.3 & 100   & 0.9  \\ 
        +CBAM & 99.3 & 100   & 0.9  \\ 
        +LGA (ours) & \textbf{99.4 (+0.1)} & 100   & 0.9  \\ 
    \hline
    \end{tabular}}
\end{table}

\textbf{Fashion-MNIST Dataset Experiment}: We further compared classification performance on the more challenging Fashion-MNIST dataset. As shown in Table \ref{tabel2}, the Local-Global Attention mechanism achieved the highest Top-1 accuracy of 92.9\%, standing out as the only attention mechanism to produce a positive gain in accuracy, while maintaining comparable computational complexity to other mechanisms.

\begin{table}[!ht] 
    \centering 
    \caption{Classification performance of different attention mechanisms integrated into the MobileNetV3 model on the Fashion-MNIST dataset \cite{xiao2017/online}.} \label{tabel2} 
    \resizebox{\linewidth}{!}{
    \begin{tabular}{lccc}
    \hline
        Backbone & Top-1 Acc (\%) & Top-5 Acc (\%) & GFLOPs  \\ \hline
        MobileNetV3 & 92.8 & 99.9 & 0.9   \\ 
        +MHSA & 92.1 & 99.9 & 0.9   \\ 
        +SE & 92.7 & 99.9 & 0.9   \\ 
        +CBAM & 92.5 & 99.9 & 0.9   \\ 
        +LGA (ours) & \textbf{92.9 (+0.1)} & 99.9 & 0.9  \\ 
    \hline
    \end{tabular}}
\end{table}

Results from these two experiments indicate that, compared to other attention mechanisms, Local-Global Attention can effectively enhance the classification performance of the model without increasing computational complexity. Whether on the simpler MNIST dataset or the more challenging Fashion-MNIST dataset, Local-Global Attention consistently demonstrated superior performance.

\subsection{Ablation Study on Attention Mechanisms}

To evaluate the effectiveness of our proposed Local-Global Attention (LGA) mechanism, we conducted ablation experiments on the TinyPerson dataset \cite{yu2020scale} using the YOLOv8 framework with default parameter settings. These experiments aim to quantify the performance contribution of Local-Global Attention, as well as individual Local Attention (LA) and Global Attention (GA), on various backbone networks. Performance is evaluated using mAP50 and mAP50-95 metrics, with experiments grouped by three backbone types: MobileNetV3 \cite{howard2019searching}, ResNet18 \cite{he2016deep}, and YOLOv8 \cite{yolov8_ultralytics}.

\textbf{MobileNetV3 Backbone}:
We integrated various attention mechanisms into the MobileNetV3  \cite{howard2019searching} backbone network. As shown in Table \ref{tabel3}, our proposed Local-Global Attention mechanism achieved an increase of 0.92 in mAP@50 and 0.29 in mAP@50-95, outperforming other attention mechanisms. These results indicate that Local-Global Attention captures both local and global dependencies more effectively than traditional attention mechanisms. Although individual GA and LA contribute to performance gains, their combination (Local-Global Attention) yields the highest accuracy.

\begin{table}[!ht]
    \centering
    \caption{Comparison of different attention mechanisms on the TinyPerson \cite{yu2020scale} dataset using MobileNetV3 \cite{howard2019searching} as the backbone.}
    \label{tabel3}
    \resizebox{\linewidth}{!}{
    \begin{tabular}{lccc}
    \hline
        Backbone & mAP50(\%) & mAP50-95(\%) & GFLOPs  \\ \hline
        MobileNetV3 & 9.88 & 3.56 & 2.8  \\ 
        +MHSA & 10.4 & 3.71 & 2.8  \\ 
        +SE & 10.1 & 3.59 & 2.8  \\ 
        +ECA & 9.96 & 3.58 & 2.8  \\ 
        +CBAM & 10.4 & 3.74 & 2.8  \\ 
        +SA & 10.3 & 3.7 & 2.8  \\ 
        +LA (ours) & 10.3 & 3.61 & 2.8  \\ 
        +GA (ours) & 10.5 & 3.81 & 2.8  \\ 
        +LGA (Ours) & \textbf{10.8 (+0.92)} & \textbf{3.85 (+0.29)} & 2.8  \\ \hline
    \end{tabular}}
\end{table}

\begin{table}[!ht]
    \centering
    \caption{Comparison of different attention mechanisms on the TinyPerson \cite{yu2020scale} dataset using ResNet18 \cite{he2016deep} as the backbone.}
    \label{tabel4}
    \resizebox{\linewidth}{!}{
    \begin{tabular}{lccc}
    \hline
        Backbone & mAP50(\%) & mAP50-95(\%) & GFLOPs  \\ \hline
        ResNet18 & 12.5 & 4.48 & 3.2  \\ 
        +MHSA & 12.4 & 4.39 & 3.3  \\ 
        +SE & 12.4 & 4.45 & 3.2  \\ 
        +ECA & 12.3 & 4.37 & 3.2  \\ 
        +CBAM & 12.6 & 4.42 & 3.2  \\ 
        +SA & 12.3 & 4.36 & 3.2  \\ 
        +LA (ours) & 12.7 & 4.53 & 3.3  \\ 
        +GA (ours) & 12.7 & 4.49 & 3.3  \\ 
        +LGA (Ours) & \textbf{12.7 (+0.2)} & \textbf{4.62 (+0.14)} & 3.3  \\ \hline
    \end{tabular}}
\end{table}

\textbf{ResNet18 Backbone}:
To assess the generalizability of Local-Global Attention, we integrated it into the ResNet18 \cite{he2016deep} backbone network. Table \ref{tabel4} shows that Local-Global Attention improved mAP@50 by 0.2 and mAP@50-95 by 0.14, achieving superior results compared to other attention mechanisms. While other mechanisms, such as SE, ECA, and CBAM, also enhanced performance, Local-Global Attention demonstrated better capabilities in capturing details and contextual information, highlighting its effectiveness in small-object detection tasks.

\textbf{YOLOv8 Backbone}:
Finally, we integrated Local-Global Attention and other attention mechanisms into the YOLOv8 \cite{yolov8_ultralytics} backbone network for the TinyPerson dataset. As shown in Table \ref{tabel5}, Local-Global Attention achieved improvements of 0.7 in mAP@50 and 0.31 in mAP@50-95, outperforming other attention mechanisms. Compared to other mechanisms, Local-Global Attention exhibited greater adaptability, effectively balancing local and global attention to enhance detection accuracy.

\begin{table}[!ht]
    \centering
    \caption{Comparison of different attention mechanisms on the TinyPerson \cite{yu2020scale} dataset using YOLOv8 \cite{yolov8_ultralytics} as the backbone.}
    \label{tabel5}
    \resizebox{\linewidth}{!}{
    \begin{tabular}{lcccccc}
    \hline
        Backbone & mAP50(\%) & mAP50-95(\%) & GFLOPs  \\ \hline
        YOLOv8 & 14.5 & 5.16 & 8.2  \\ 
        +MHSA & 14.6 & 5.22 & 8.4  \\ 
        +SE & 15.2 & 5.41 & 8.2  \\ 
        +ECA & 15.1 & 5.39 & 8.2  \\ 
        +CBAM & 14.7 & 5.31 & 8.2  \\ 
        +SA & 15 & 5.36 & 8.2  \\ 
        +LA (ours) & 15.1 & 5.34 & 8.5  \\ 
        +GA (ours) & 14.9 & 5.36 & 8.5  \\ 
        +LGA (Ours) & \textbf{15.2 (+0.7)} & \textbf{5.47 (+0.31)} & 8.5 \\ \hline
    \end{tabular}} 
\end{table}

Results across these three backbone networks indicate that, while individual Local Attention and Global Attention improve model performance, the combined Local-Global Attention mechanism consistently yields the best results. By dynamically adjusting the balance between local and global features, Local-Global Attention significantly enhances target detection, making it a robust choice for handling complex detection tasks in constrained environments.

\subsection{Applications}

In this section, we evaluate the effectiveness of the proposed Local-Global Attention mechanism across various vision tasks by conducting experiments on multiple datasets. By comparing  Local-Global Attention with commonly used attention mechanisms, we assess its performance on diverse benchmark datasets. The experiments encompass object detection tasks across different dataset scenarios to explore  Local-Global Attention's applicability and transferability.

\subsubsection{Object Detection Experiments}

\noindent\textbf{Experimental Setup}: For this set of experiments, we used the VisDrone2019, VOC2012, and COCOminitrain datasets. All models were based on the MobileNetV3 architecture, optimized with SGD \cite{ruder2016overview}, and utilized YOLOv8 as the detection framework. We evaluated each attention mechanism's impact on model detection performance using mAP50 and mAP50-95 metrics.

\noindent\textbf{Results}: As shown in Table \ref{tabel6}, Local-Global Attention achieved the best performance on most metrics compared to other attention mechanisms. On the VisDrone2019 dataset, Local-Global Attention increased the mAP@50-95 score to 11.5, representing a 0.3 improvement over the baseline. For the VOC2012 dataset, Local-Global Attention increased mAP@50 by 0.1 and mAP@50-95 by 0.7. In the COCOminitrain dataset, Local-Global Attention also achieved the best results across all evaluation metrics. Overall, Local-Global Attention demonstrated superior capability in capturing detailed features and extracting global information, resulting in higher accuracy across multiple detection tasks.

\begin{table*}
    \centering
    \caption{The experimental results of different attention mechanisms on the VOC2012, VisDrone2019, and COCOminitrain datasets.  }
    \label{tabel6}
    \tabcolsep=0.3cm
    \resizebox{\linewidth}{!}{
    \begin{tabular}{cclcccc}
    \hline
        datasets & Method & Backbone & mAP50(\%) & mAP50-95(\%) & Parameters(M) & GFLOPs  \\ \hline
        VOC2012 & YOLOv8 & MobileNetV3 & 43.7 & 27.5 & 1.19 & 2.8  \\ 
        ~ & YOLOv8 & MobileNetV3+MHSA & 37.4 & 22.9 & 1.19 & 2.8  \\ 
        ~ & YOLOv8 & MobileNetV3+SE & 42.6 & 27.0 & 1.19 & 2.8  \\ 
        ~ & YOLOv8 & MobileNetV3+CBAM & 41.9 & 26.6 & 1.19 & 2.8  \\ 
        ~ & YOLOv8 & MobileNetV3+LGA & \textbf{43.8 (+0.1)} & \textbf{28.2 (+0.7)} & 1.21 & 2.8  \\ \hline

        VisDrone2019 & YOLOv8 & MobileNetV3 & 20.9 & 11.2 & 1.19  & 2.8  \\ 
        ~ & YOLOv8 & MobileNetV3+MHSA & 20.9 & 11.4 & 1.19  & 2.8  \\ 
        ~ & YOLOv8 & MobileNetV3+SE & 20.7 & 11.3 & 1.19  & 2.8  \\ 
        ~ & YOLOv8 & MobileNetV3+CBAM & 21.2 & \textbf{11.6} & 1.19  & 2.8  \\ 
        ~ & YOLOv8 & MobileNetV3+LGA & \textbf{21.3 (+0.4)} & 11.5 (+0.3) & 1.19  & 2.8  \\ \hline

        COCOminitrain & YOLOv8 & MobileNetV3 & 14.6 & 8.8 & 1.42  & 4.0  \\ 
        ~ & YOLOv8 & MobileNetV3+MHSA & 11.7 & 6.86 & 1.45  & 4.0  \\ 
        ~ & YOLOv8 & MobileNetV3+SE & 15.0 & 9.03 & 1.45  & 4.0  \\ 
        ~ & YOLOv8 & MobileNetV3+CBAM & 13.8 & 8.15 & 1.45  & 4.0  \\ 
        ~ & YOLOv8 & MobileNetV3+LGA & \textbf{15.1 (+0.5)} & \textbf{9.08 (+0.28)} & 1.47  & 4.0 \\ \hline
    \end{tabular}}
\end{table*}

\begin{table*}
    \centering
    \caption{The experimental results of different attention mechanisms on the DOTAv1.0 and GWHD2020 datasets. }
    \label{tabel7}
    \tabcolsep=0.3cm
    \resizebox{\linewidth}{!}{
    \begin{tabular}{cclcccc}
    \hline
        datasets & Method & Backbone & mAP50(\%) & mAP50-95(\%) & Parameters(M) & GFLOPs  \\ \hline
        DOTAv1.0 & YOLOv8 & MobileNetV3 & 53.9 & 32.4 & 1.19  & 2.8  \\ 
        ~ & YOLOv8 & MobileNetV3+MHSA & 53.6 & 32.4 & 1.19  & 2.8  \\ 
        ~ & YOLOv8 & MobileNetV3+SE & \textbf{54.4} & 32.7 & 1.19  & 2.8  \\ 
        ~ & YOLOv8 & MobileNetV3+CBAM & 53.7 & 32.4 & 1.19  & 2.8  \\ 
        ~ & YOLOv8 & MobileNetV3+LGA & 54.3 (+0.4) & \textbf{32.8 (+0.4)} & 1.21  & 2.8  \\ \hline

        GWHD2020 & YOLOv8 & MobileNetV3 & 95.5 & 60 & 1.19  & 2.8  \\ 
        ~ & YOLOv8 & MobileNetV3+MHSA & 95.6 & 59.5 & 1.19  & 2.8  \\ 
        ~ & YOLOv8 & MobileNetV3+SE & 95.6 & 59.6 & 1.19  & 2.8  \\ 
        ~ & YOLOv8 & MobileNetV3+CBAM & 95.7 & 60 & 1.19  & 2.8  \\ 
        ~ & YOLOv8 & MobileNetV3+LGA(ours) & \textbf{95.8 (+0.3)} & \textbf{60.1 (+0.1)} & 1.20  & 2.8  \\ \hline
    \end{tabular}}
\end{table*}

\subsubsection{Extended Dataset Experiments}

\noindent\textbf{Experimental Setup}: In this set of experiments, we applied the models to the DOTAv1.0 and GWHD2020 datasets to evaluate performance in complex scenes and high-resolution images. The base model remained MobileNetV3, with the optimizer switched to Adam \cite{ruder2016overview}, and YOLOv8’s default parameter settings were maintained.

\noindent\textbf{Results}: Table \ref{tabel7} presents the results of different attention mechanisms on the DOTAv1.0 and GWHD2020 datasets. On the DOTAv1.0 dataset, Local-Global Attention achieved the highest mAP@50-95 score of 32.8. On the GWHD2020 dataset, Local-Global Attention also yielded the best performance, reaching 95.8 for mAP@50 and 60.1 for mAP@50-95, representing increases of 0.3 and 0.1 over the baseline, respectively. These results demonstrate that Local-Global Attention maintains high detection performance in more complex scenarios and achieves robust transferability across different datasets.

The above experiments demonstrate that the local-global attention mechanism exhibits outstanding performance across different detection tasks and datasets, highlighting its superior performance and broad applicability in visual tasks.

\section{Conclusions}
In summary, this paper introduces a novel Local-Global Attention mechanism that enhances feature representation by dynamically balancing fine-grained local features with broader global context. This enables the model to capture detailed local information while maintaining an understanding of the broader global context. The Local-Global Attention mechanism also incorporates learnable alpha parameters, which adaptively adjust the weights of local and global features, and further improves spatial understanding through positional encoding.

Experimental evaluations on several benchmark datasets show that the Local-Global Attention mechanism performs well in standard scenarios, and demonstrates remarkable flexibility and robustness in more complex cases, where a fine balance between local details and global context is required, consistently outperforming traditional attention mechanisms. Moreover, it maintains high computational efficiency, offering a clear advantage in resource-constrained applications. These experimental results confirm that our approach not only improves model performance but also effectively addresses the challenges faced by existing attention mechanisms in capturing both fine-grained details and large-scale spatial structures.

\clearpage

{
    \small
    \bibliography{main.bb}
}
\clearpage

\end{document}